\pdfoutput=1

\documentclass[11pt]{article}

\usepackage[final]{acl}

\usepackage{times}
\usepackage{latexsym}

\usepackage[T1]{fontenc}
\usepackage{amsfonts}
\usepackage[utf8]{inputenc}

\usepackage{microtype}

\usepackage{inconsolata}

\usepackage{graphicx}
\usepackage[table]{xcolor}
\usepackage{enumitem}
\usepackage{amsmath}
\usepackage{bbm}
\usepackage{appendix}
\usepackage{multirow}
\usepackage{booktabs}
\usepackage{makecell}
\usepackage[normalem]{ulem}
\useunder{\uline}{\ul}{}

\newcommand{\cf}{\emph{cf. }}

\title{Assistant-Guided Mitigation of Teacher Preference Bias in LLM-as-a-Judge}

\renewcommand*{\thefootnote}{\fnsymbol{footnote}}
\author{
Zhuo Liu\textsuperscript{1},
~Moxin Li\textsuperscript{2,$\dagger$},
~Xun Deng\textsuperscript{1}, 
\textbf{Qifan Wang}\textsuperscript{3},
~\textbf{Fuli Feng}\textsuperscript{1,$\dagger$}, 
\\
\textsuperscript{1}University of Science and Technology of China
~\textsuperscript{2}National University of Singapore
~\textsuperscript{3}Meta AI,
\\
\tt{liuz\_@mail.ustc.edu.cn},~\tt{limoxin@u.nus.edu},~\tt{dx981228@mail.ustc.edu.cn},
\\\tt{wqfcr@fb.com},~\tt{fulifeng93@gmail.com}
}

\begin{document} 
\maketitle
\footnotetext{\textsuperscript{$\dagger$}Corresponding author.}
\renewcommand*{\thefootnote}{\arabic{footnote}}

\begin{abstract}
LLM-as-a-Judge employs large language models (LLMs), such as GPT-4, to evaluate the quality of LLM-generated responses, gaining popularity for its cost-effectiveness and strong alignment with human evaluations.
However, training proxy judge models using evaluation data generated by powerful teacher models introduces a critical yet previously overlooked issue: teacher preference bias, where the proxy judge model learns a biased preference for responses from the teacher model. 
To tackle this problem, we propose a novel setting that incorporates an additional assistant model, which is not biased toward the teacher model's responses, to complement the training data. 
Building on this setup, we introduce AGDe-Judge, a three-stage framework designed to debias from both the labels and feedbacks in the training data. Extensive experiments demonstrate that AGDe-Judge effectively reduces teacher preference bias while maintaining strong performance across six evaluation benchmarks\footnote{Code is available at \url{https://github.com/Liuz233/AGDe-Judge}.}.
\end{abstract}

\section{Introduction}

LLM-as-a-Judge refers to the use of Large Language Models (LLMs), such as GPT-4 \cite{openai2024gpt4technicalreport}, to evaluate text quality by generating feedback and making evaluative judgments \cite{zheng2023judging}. 
Unlike traditional manual evaluation or automatic metrics (e.g., BLEU), this method offers a cost-effective and scalable alternative, achieving strong alignment with human evaluations when using advanced models like GPT-4 \cite{liu2023g}.
Consequently, LLM-as-a-Judge has seen growing adoption in LLM evaluation tasks \cite{zheng2023judging,wu2024meta,dubois2025lengthcontrolledalpacaevalsimpleway}.

Concerns regarding the high cost, limited transparency, and lack of controllability of proprietary large models (e.g., GPT-4) have driven the adoption of proxy judge models based on open-source LLMs \cite{wangpandalm, kim2023prometheus, kim2024prometheus, ligenerative}. To equip smaller open-source models with evaluation capabilities comparable to those of advanced large models, this approach typically employs a high-performing teacher model—such as GPT-4—to generate judge-specific training data for evaluation tasks. The open-source model is then fine-tuned on this data to create a specialized proxy judge model. By constructing high-quality judge datasets and applying techniques such as weight merging \cite{kim2024prometheus}, the resulting proxy judge models can achieve performance close to that of large models like GPT-4 \cite{kim2023prometheus, ligenerative}, enabling broad practical adoption.

\begin{figure}[t]
    \centering
    \setlength{\abovecaptionskip}{0.05cm}
    \setlength{\belowcaptionskip}{0cm}
    \includegraphics[width=\linewidth]{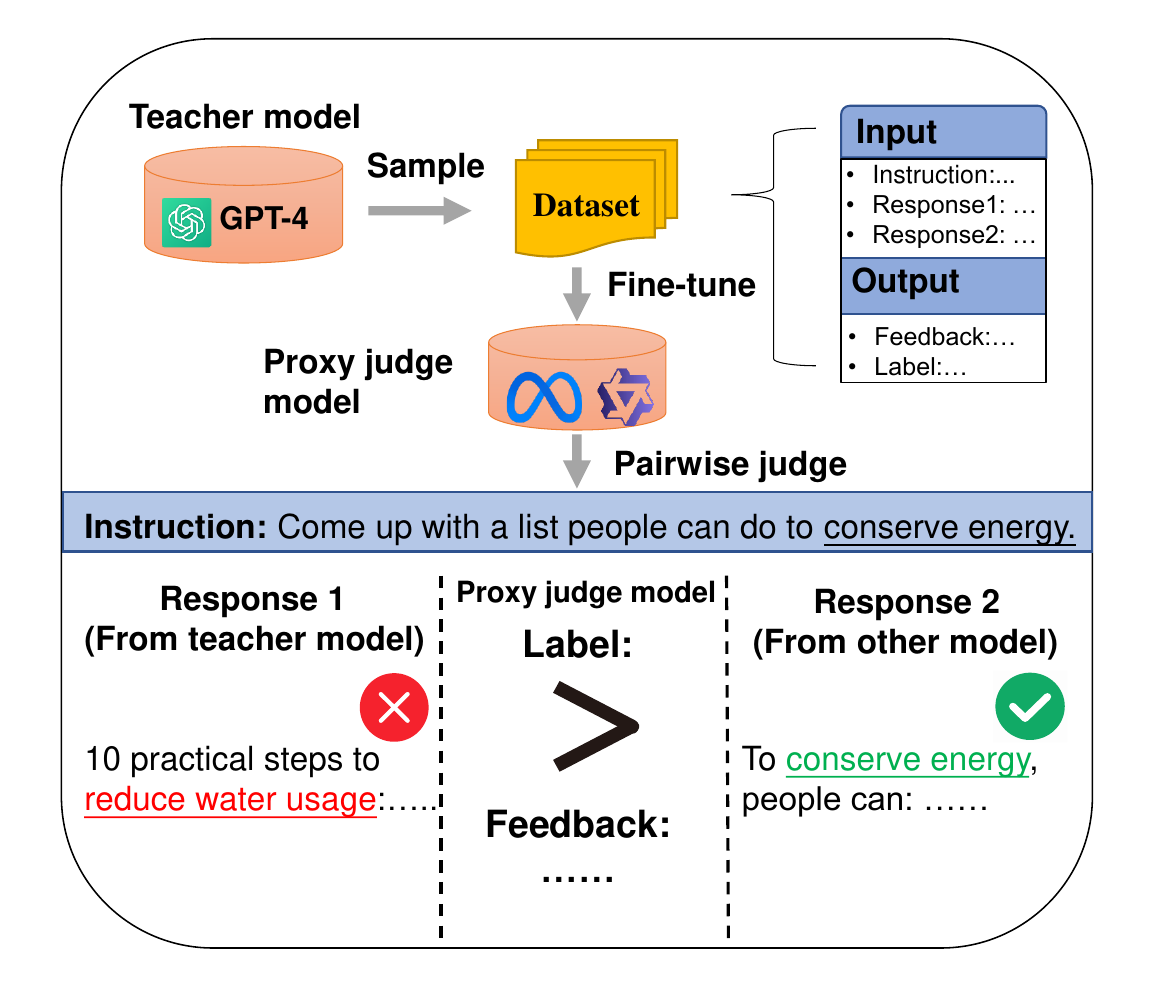}
    \caption{Illustration of the teacher preference bias. The evaluation task is a pairwise ranking, where the proxy judge model generates evaluation feedback and a label indicating which of the two responses is better.}
    \label{fig:intro}
    \vspace{-15pt}
\end{figure}

Despite the advantages of proxy judge models, we identify a key limitation: these models exhibit a significant bias favoring responses generated by the teacher model, regardless of their actual quality. We term this phenomenon teacher preference bias (\cf Figure~\ref{fig:intro}). This bias stems from the self-preference bias of the teacher model itself~\cite{ye2024justice, chen2025llm}, which is contained in the training data and subsequently captured by the proxy judge model through fine-tuning. To the best of our knowledge, teacher preference bias has not been previously identified or systematically investigated. Our experiments confirm the presence of this bias, which arises from both the judge labels and feedback included in the training data. As GPT-4 is a dominant and highly capable model, many existing methods rely exclusively on data generated by GPT-4~\cite{ligenerative, kim2023prometheus, kim2024prometheus}, further amplifying the spread of this bias and posing a substantial challenge to the fairness and accuracy of proxy judge models.

To mitigate teacher preference bias, we propose to introduce an additional, smaller LLM, termed as assistant model, into the training pipeline of proxy judge models. These models do not have a biased preference toward the teacher model's responses, and are relatively inexpensive with decent evaluation capability. Within this framework, we leverage the assistant model to debias the teacher model-generated data from two sources: biased labels and biased feedback. Firstly, we aim to identify and filter out the instances with biased labels leveraging information from the reliable labels agreed by the assistant model. Then, we aim to debias the feedback by reducing the teacher model's tendency to overemphasize minor or superficial issues.

To this end, we propose a three-stage framework, AGDe-Judge (Assistant-Guided Debiasing for Judge Models), to mitigate teacher preference bias in proxy judge models.
First, we filter out biased labels using an implicit reward margin \cite{rafailov2024direct}, which is derived from consensus labels between the teacher and assistant models. 
Second, we leverage the assistant model to identify severe flaws in responses, thereby counteracting the overemphasis on minor or superficial issues in the feedback.
Finally, we fine-tune the student model using the refined data to obtain the final proxy judge model. 
Experimental results show that AGDe-Judge effectively reduces teacher preference bias while maintaining high evaluation accuracy across six standard benchmarks. 
In summary, the main contributions of our paper are as follows:
\begin{itemize}[leftmargin=*]
    \item We identify and define a previously unstudied critical limitation in existing proxy judge models, the teacher preference bias. 
    \item We introduce an assistant LLM to tackle this bias and propose a three-stage approach that tackles biased labels and biased feedback.
    \item We conduct extensive experiments validating the strong effectiveness of our approach both in debiasing and evaluation performance. 
\end{itemize}

\section{Related Work}


\textbf{LLM-as-a-Judge}
\citet{zheng2023judging} proposed LLM-as-a-Judge, leverages powerful LLMs (e.g., GPT-4) to evaluate responses to open-ended questions, offering a scalable alternative to costly human annotators and limited traditional methods. Studies show LLMs achieve high agreement with human experts \cite{ashktorab2024aligning, bavaresco2024llms}, driving its adoption across various tasks \cite{zhu2023judgelm, cui2023ultrafeedback, bai2023benchmarking}. However, this paradigm suffers from biases, including position bias \cite{shi2025judgingjudgessystematicstudy}, verbosity bias \cite{chen2024humans}, and self-preference bias, where LLMs favor their own responses \cite{liprd, panickssery2024llm}. These biases undermine the reliability and fairness of LLM-as-a-Judge evaluations \cite{ye2024justice}.

\noindent\textbf{Proxy Judge Models}
Due to the high cost, limited transparency, and lack of controllability associated with proprietary large models, fine-tuning open-source models to serve as judge models—commonly referred to as proxy judge models—has become an increasingly popular alternative~\cite{wangpandalm,zhu2023judgelm,kim2023prometheus,kim2024prometheus}. This approach typically involves leveraging a powerful proprietary model (mostly GPT-4), known as the teacher model, to construct evaluation-specific training data. However, the self-preference bias implicitly embedded in the teacher model’s outputs may be inherited by the proxy judge model, introducing subtler forms of bias. Despite its significance, this issue has not yet been systematically studied.

\noindent\textbf{Debiasing Approaches}
Various debiasing approaches have been proposed to address the widespread biases in LLM-as-a-Judge. Discussion-based methods mitigate the bias of a single LLM by leveraging multiple LLMs to engage in debate and deliberation~\cite{khan2024debating,liprd}. Adversarial methods introduce structured rationale pairs~\cite{ye2024beyond}, or construct specially curated debiasing datasets~\cite{park2024offsetbias}. However, addressing the previously unstudied teacher preference bias calls for new debiasing methods specifically tailored to this issue.
\section{Preliminary Study}
\label{sec:preliminary}

\begin{table*}[t]
\setlength{\abovecaptionskip}{0.05cm}
\setlength{\belowcaptionskip}{0cm}
\renewcommand*{\arraystretch}{1}
\centering
\resizebox{\linewidth}{!}{
\begin{tabular}{lc|ccc|ccc}
\toprule
\multirow{2}{*}{Dataset Filtering}      & \multirow{2}{*}{Proxy Judge Model} & \multicolumn{3}{c|}{AutoJ-pairwise}              & \multicolumn{3}{c}{Preference Collection}        \\
                                       &                   & MT-Bench       & Arena-Human    & OffsetBias     & MT-Bench       & Arena-Human    & OffsetBias     \\ \hline
\multirow{2}{*}{No Filtering}        & GPT3.5-Mistral-7B & 0.695          & 0.565          & \textbf{0.269} & 0.681          & 0.538          & 0.314 \\
                                       & GPT4-Mistral-7B   & \begin{tabular}[c]{@{}c@{}}{\textcolor{red}{(+0.7\%)}}\\ 0.700\end{tabular} & \begin{tabular}[c]{@{}c@{}}{\textcolor{red}{(+3.2\%)}}\\ 0.583\end{tabular} & \begin{tabular}[c]{@{}c@{}}{\textcolor{blue}{(-32.3\%)}}\\ 0.182\end{tabular}         & \begin{tabular}[c]{@{}c@{}}{\textcolor{red}{(+7.8\%)}}\\ \textbf{0.734}\end{tabular} & \begin{tabular}[c]{@{}c@{}}{\textcolor{red}{(+8.7\%)}}\\ \textbf{0.585}\end{tabular} & \begin{tabular}[c]{@{}c@{}}{\textcolor{blue}{(-19.4\%)}}\\ 0.253\end{tabular}          \\ \hline
\multirow{2}{*}{Same-Label Filtering} & GPT3.5-Mistral-7B & 0.712          & 0.563          & 0.230 & 0.696          & 0.533          & \textbf{0.320} \\
                                       & GPT4-Mistral-7B   & \begin{tabular}[c]{@{}c@{}}{\textcolor{red}{(+2.2\%)}}\\ \textbf{0.728}\end{tabular} & \begin{tabular}[c]{@{}c@{}}{\textcolor{red}{(+4.8\%)}}\\ \textbf{0.590}\end{tabular} & \begin{tabular}[c]{@{}c@{}}{\textcolor{blue}{(-21.7\%)}}\\ 0.180\end{tabular}          & \begin{tabular}[c]{@{}c@{}}{\textcolor{red}{(+4.2\%)}}\\ 0.725\end{tabular} & \begin{tabular}[c]{@{}c@{}}{\textcolor{red}{(+5.4\%)}}\\ 0.562\end{tabular} & \begin{tabular}[c]{@{}c@{}}{\textcolor{blue}{(-22.8\%)}}\\ 0.247\end{tabular}          \\ \bottomrule
\end{tabular}
}
\caption{Evaluation accuracy of Mistral-7B-Instruct-v0.3 fine-tuned on the AutoJ-pairwise and Preference Collection
datasets under different filtering condition. The best accuracy for each benchmark is \textbf{bolded}.}
\label{tb:preliminary_t}
\vspace{-8pt}
\end{table*}

\paragraph{Problem Definition}
\label{sec:problem definition}
LLM-as-a-judge is an efficient and effective strategy for evaluating LLM-generated responses to compare LLM performance.  
We focus on the most common and effective evaluation setting of \emph{pairwise ranking} \cite{kim2024prometheus}, where the judge LLM evaluates and rank pairs of LLM responses. 
Let $M$ denote the judge LLM. 
Given an input tuple consisting of a instruction $q$, two responses $r_0$ and $r_1$ generated by LLMs to be evaluated, and an auxiliary input $e$ which includes evaluation criteria and reference answers~\cite{kim2024prometheus}, the judge LLM generates a textual feedback $v$ to illustrate the reason behind its judgment, and a preference label $y \in \{0, 1\}$ to indicate which response is better. Formally, 
\begin{align}
    (v, y) = M(q, r_0, r_1, e). \label{eq:definition_llm_as_a_judge}
\end{align} 

Due to the limitations of proprietary large judge models (e.g., GPT-4), such as high cost, limited transparency and controllability, researchers have increasingly turned to proxy judge models as a practical alternative. A common approach begins by constructing a training dataset using outputs from the proprietary model, referred to as the teacher model $M_t$, denoted as $\mathcal{D}_{t} = \{(q, r_0, r_1, e), (v_t, y_t)\}$, where $v_t$ and $y_t$ are the textual feedback and label generated by $M_t$. An open-source model is then fine-tuned on $\mathcal{D}_{t}$ to obtain a proxy judge model, referred to as the student model $M_s$, which is used for evaluation tasks. This method of constructing proxy judge models has demonstrated strong performance comparable to that of the teacher model \cite{kim2024prometheus}. Formally, denoting the feedback and label generated from proxy judge model as $v_s$ and $y_s$, 
\begin{align}
    (v_s, y_s) = M_s(q, r_0, r_1, e). \label{eq:definition_proxy_judge_model}
\end{align}

\paragraph{Teacher Preference Bias}
Despite the advantages of proxy judge models, we identify a critical issue in pairwise ranking: when \emph{one response is generated by the teacher model} and the other by an unrelated model, the proxy judge often exhibits a \emph{biased preference for the teacher model’s response}, which is much larger than the actual probability of the teacher model's response being better . We refer to this phenomenon as teacher preference bias. This bias stems from the self-preference bias~\cite{ye2024justice, chen2025llm} captured in $\mathcal{D}_t$ generated by the teacher model who often favors its own responses. Denoting the ground-truth label as $y_{gt} \in \{0, 1\}$, and the response index generated by $M_t$ as $y_{tg} \in \{0, 1\}$, this bias can exhibit as
\begin{align}
    P(y_s = y_{tg}) \gg P(y_{gt} = y_{tg}). \label{eq:definition_teacher_preference_bias}
\end{align}

This issue poses a significant challenge to building reliable proxy judge models. First, existing evaluation datasets (e.g., Preference Collection~\cite{kim2024prometheus}, AutoJ-pairwise~\cite{ligenerative}, JudgeLM~\cite{zhu2023judgelm}) heavily rely on the state-of-the-art teacher model GPT-4, which is widely used due to its strong alignment with human preferences \cite{liu2023g}. Moreover, the response pairs in these datasets often include outputs generated by GPT-4, which creates the condition for introducing such bias. Also, many evaluation benchmarks involve direct comparisons between a GPT-4 response and a response from another model (e.g., MT-Bench~\cite{zheng2023judging}, Chatbot-Arena~\cite{chiang2024chatbot}, Reward-Bench~\cite{lambert2024rewardbench}). As a result, the proxy judge model can easily learn and exhibit this latent bias.
In the following section, we systematically demonstrate the existence of this bias.

\begin{figure*}[t]
    \centering
    \setlength{\abovecaptionskip}{0.05cm}
    \setlength{\belowcaptionskip}{0cm}
    \scalebox{1}[0.85]{\includegraphics[width=\linewidth]{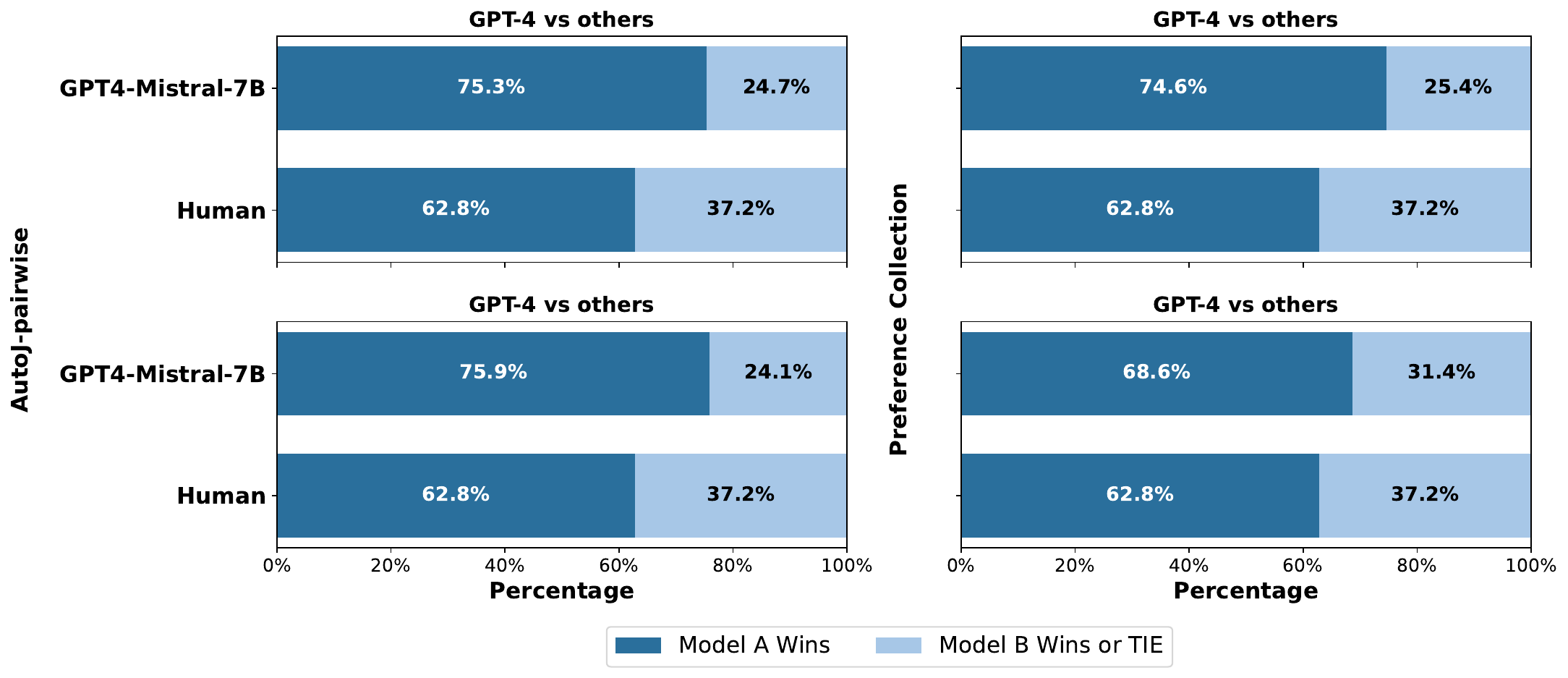}}
    \caption{Results of GPT4-Mistral-7B and Human judgments on MT-Bench Human Judge (filtered). The models were trained on AutoJ-pairwise (left) and Preference Collection data (right). The upper row uses all training samples; the lower row uses only those where GPT-4 and GPT-3.5-Turbo assign the same preference label.}
    \label{fig:preliminary}
    \vspace{-15pt}
\end{figure*}

\paragraph{Evaluation Setup}
\label{sec:preliminary_setup}
To demonstrate the existence of teacher preference bias, we fine-tune proxy judge models using GPT-4 as the teacher model and evaluate their biased preference toward GPT-4-generated responses. The \textbf{bias evaluation} is primarily conducted on the following datasets:
\begin{itemize}[leftmargin=*]
    \item \textbf{OffsetBias} (filtered)~\cite{park2024offsetbias}: A representative evaluation dataset on the bias for LLM-as-a-Judge. The worse responses are generated by GPT-4, crafted intentionally through adversarial prompting techniques such as providing misleading or off-topic instructions, or explicitly encouraging the inclusion of factual errors or incomplete answers. In contrast, the better responses come from other models. Lower accuracy in this dataset show more severe bias. 
    \item \textbf{MT-Bench Human Judge} (filtered)~\cite{zheng2023judging}: Uses the 80 prompts from MT-Bench along with 3.3k pairs of model responses annotated with human preferences.
    We extract a subset of MT-Bench Human Judge consisting of instances where one response is from GPT-4 and the other from a different model. To evaluate the bias, we measure the extent to which the proxy judge favors GPT-4 responses compared to the ground-truth human labels. 
\end{itemize}

Additionally, we employ two \textbf{standard benchmarks}, the full \textbf{MT-Bench Human Judge} and \textbf{Arena-Human}~\cite{chiang2024chatbot}, from which we randomly sample evaluation instances to examine the quality of the proxy judge model.

For \textbf{fine-tuning datasets}, we employ \textbf{AutoJ-pairwise} and \textbf{Preference Collection}. Both are constructed under GPT-4 with evaluated responses from GPT-4. 
In addition to using GPT-4 as the teacher model, we investigate the severity of teacher preference bias by introducing \textbf{an alternative teacher model: GPT-3.5-Turbo}. We prompt GPT-3.5-Turbo with the same inputs used in the fine-tuning datasets to generate feedback and labels. Comparing the performance of proxy judge models fine-tuned with GPT-3.5-Turbo versus GPT-4 as teacher enables us to evaluate the extent of bias toward GPT-4-generated responses.
For \textbf{base proxy judge model}, we use the open-source \textbf{Mistral-7B-Instruct-v0.3}~\cite{jiang2023mistral7b}. 

Furthermore, we investigate \textbf{different sources of bias}. We hypothesize that bias may arise from both the preference labels and the textual feedback. To isolate the effect of feedback, we introduce an additional setting where we retain only training instances for which GPT-4 and GPT-3.5-Turbo teacher models assign identical preference labels. This allows us to assess the extent of bias attributable specifically to the feedback. More details regarding dataset processing, fine-tuning, and generating procedures are in Appendix~\ref{appdA}.

\paragraph{Evaluation Results}
Table~\ref{tb:preliminary_t} shows the bias and accuracy results for proxy judge models with different teacher models. 
On standard benchmarks, the model trained with GPT-4 as the teacher (GPT4-Mistral-7B) achieves the highest accuracy, outperforming GPT3.5-Mistral-7B. This highlights GPT-4’s strong alignment with human evaluations and supports its use as a teacher model for generating fine-tuning data. 
However, on the bias evaluation set OffsetBias, GPT4-Mistral-7B shows a notable performance drop compared to GPT3.5-Mistral-7B. This suggests that GPT4-Mistral-7B incorrectly favors GPT-4 responses even when they are worse. Together, these findings indicate that, despite its overall effectiveness, the proxy judge model tends to over-rank responses from its teacher model—a phenomenon we refer to as \emph{teacher preference bias}.

Results using identical labels from both teacher models are also presented below Table~\ref{tb:preliminary_t}, showing a trend consistent with the unfiltered results: while GPT4-Mistral-7B achieves higher accuracy on standard benchmarks, it still demonstrates a clear biased preference toward responses from GPT-4, confirming the presence of teacher preference bias. Notably, the bias performance gap in the feedback-only setting is smaller than that in the joint setting. Taken together, these findings indicate that both labels and feedback in the training data contribute to the teacher preference bias.

Figure~\ref{fig:preliminary} presents the results on MT-Bench Human Judge (filtered), offering an complementary metric, namely by comparing the proportion of GPT-4 responses labeled as winner by judge models with the actual proportion labeled as winner by human annotators, thereby quantifying the degree of teacher preference bias. We can clearly observe that, under different training datasets and sampling settings, the win ratio assigned to GPT-4 by GPT4-Mistral-7B is substantially higher than that assigned by human annotators. This provides another piece of intuitive evidence that reinforces the existence of teacher preference bias.

\paragraph{Case Study}
To further investigate the causes of teacher preference bias, we conduct a case study. We first observe the patterns of GPT-4’s self-preference bias in the training data, as illustrated in Table~\ref{case_preliminary_autoj}. 
The case reveals that GPT-4’s feedback is not solely grounded in core evaluation criteria such as relevance, completeness, and factual accuracy. Instead, it also emphasizes features commonly associated with GPT-4’s own outputs—such as informativeness, extensive citations, and greater depth of reasoning.
Next, we examine the failure patterns of proxy judge models with respect to teacher preference bias, as shown in Table~\ref{case_preliminary_offsetbias}. We find that the model tends to prioritize stylistic and content-rich features like vividness and depth of knowledge—features typical of GPT-4 outputs—while overlooking critical flaws such as topical irrelevance. This case shows that self-preference bias present in the training data is captured by the proxy judge model during fine-tuning.


\section{Methodology}
\begin{figure*}[t]
    \centering
    \setlength{\abovecaptionskip}{0.05cm}
    \setlength{\belowcaptionskip}{0cm}
    \includegraphics[width=\linewidth]{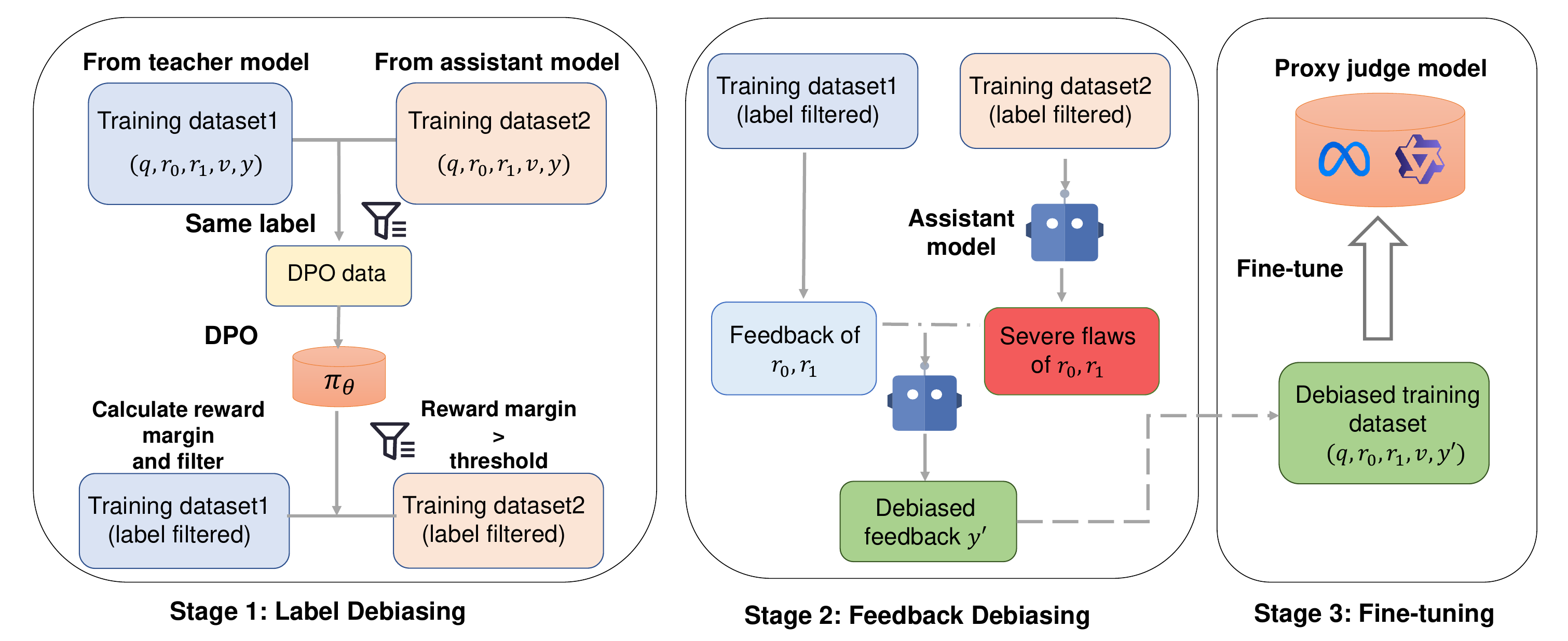}
    \caption{Illustration of three-stage debiasing framework AGDe-Judge. $q$, $r_0$, $r_1$, $v$, $y$ denote the inputs and outputs of the pairwise ranking task, following the same notation as in Section~\ref{sec:problem definition}.}
    \label{fig:method2}
    \vspace{-15pt}
\end{figure*}

After identifying the teacher preference bias, our goal is to mitigate this bias in proxy judge models while maintaining their strong evaluation performance. We incorporate an additional smaller LLM, referred to as the assistant model, into the fine-tuning process. The assistant model is cost-effective, exhibits reliable evaluation capabilities, and remains unaffected by the teacher model's bias.
Building on our earlier observations, the proposed framework aims to reduce teacher preference bias by targeting two key sources sequentially: first, by filtering biased labels, and then by addressing biased feedback. We propose AGDe-Judge, a three-stage debiasing framework to achieve this goal.


\paragraph{Stage 1: Label Filtering by Reward Margin}
We observe that labels in the dataset $\mathcal{D}_t$ can be incorrect and biased. Therefore, the first stage of our framework focuses on filtering out incorrectly labeled instances. Inspired by \citet{DBLP:journals/corr/abs-2502-14560}, we assign a reward score to each candidate response in an evaluation instance and retain only those instances with a sufficiently large reward margin between the better and worse responses—indicating a clear quality difference. To ensure the reward is reliable and corrects potential bias from the teacher model, we leverage instances where the teacher and assistant models agree on the ranking, rather than relying solely on the teacher model’s judgment.

Specifically, we first construct an additional training dataset by prompting the assistant model with the same instructions and candidate responses, generating new labels and feedback. We then select samples where both the teacher and assistant models provide consistent label annotations, denoted as $\mathcal{D}_{a}$. These samples are used to train an auxiliary DPO model $M_i$, using the following loss function:
\begin{align}
    \mathcal{L}_i = -\mathbb{E}_{\mathcal{D}_a}\bigg[\log \sigma\bigg(&\beta\frac{M_i(r_y|q)}{M_{\text{ref}}(r_{y}|q)}\nonumber \\
    & - \beta\frac{M_i(r_{1-y}|q)}{M_{\text{ref}}(r_{1-y}|q)}\bigg)\bigg],
\end{align}
where $M_{\text{ref}}$ is the reference assistant model. 
Next, we compute the implicit reward margin between the better and worse responses, $r_y$ and $r_{1-y}$:
\begin{align}
    t = \log \frac{M_i(r_y|q)}{M_{\text{ref}}(r_{y}|q)} - \log \frac{M_i(r_{1-y}|q)}{M_{\text{ref}}(r_{1-y}|q)}.
\end{align}
By applying a threshold $T$ on this reward margin, we filter out instances with potentially biased labels in both datasets. The remaining better-quality samples are then passed to the next stage.

\paragraph{Stage 2: Feedback Debiasing by Assistant-Critique Aggregation}

For instances with filtered labels, the feedback provided by the teacher model may still exhibit significant bias—overemphasizing minor features characteristic of its own responses while overlooking more critical aspects of response quality (see Section~\ref{sec:preliminary}). In contrast, the assistant model’s feedback is less prone to such self-preference bias. 

To address this, we utilize the assistant model to identify severe flaws in both candidate responses. These identified flaws serve as anchors to guide the assistant model in mitigating the biased focus of the original feedback on minor or superficial features. The assistant model is then prompted to generate revised feedback by integrating the detected flaws with the original teacher feedback, resulting in a more balanced and critical evaluation. Formally, 
\begin{align}
    c_a &= M_a(q, r_0, r_1, w), \\
    v_a &= M_a(q, r_0, r_1, v, c_a, e, o),
\end{align}
where $M_a$ is assistant model, $c_a$ is the critique under prompt $w$, and $v_a$ is the revised feedback under original feedback $v$, $c_a$, and prompt $o$. 

\paragraph{Stage 3: Fine-tuning}
After the first two stages, we obtain de-biased labels and feedback. We then fine-tune the proxy judge model $M_s$ using the following loss function. Denoting the dataset obtained from the first two stages as $\mathcal{D}_j = \{q, r_0, r_1, e, v_a, y_a\}:$
\begin{align}
    \mathcal{L}_j = -\mathbb{E}_{\mathcal{D}_j} \bigg[ \log \frac{M_s(v_a, y_a|q, r_0, r_1, e)}{M_{\text{ref}}(v_a, y_a|q, r_0, r_1, e)} \bigg]. 
\end{align}
\section{Experiments}

\begin{table*}[t]
\setlength{\abovecaptionskip}{0.05cm}
\setlength{\belowcaptionskip}{0cm}
\renewcommand*{\arraystretch}{1}
\centering
\resizebox{\linewidth}{!}{
\begin{tabular}{cccccccc}
\toprule
\rowcolor{gray!30}
 & \textbf{Preference-Bench} & \textbf{MT-Bench} & \textbf{Reward-Bench} & \textbf{Arena-Human} & \textbf{UltraFeedback} & \textbf{JudgeLM} & \textbf{OffsetBias} \\ 
Mistral-7B                 & 0.669                     & 0.571             & 0.835                 & 0.462                & 0.554                  & 0.616            & {\ul 0.322}         \\
GPT3.5-Mistral-7B          & \textbf{0.874}            & 0.695             & 0.931                 & 0.565                & 0.681                  & 0.735            & 0.269               \\
GPT4-Mistral-7B            & 0.721                     & 0.700             & {\ul 0.962}           & 0.583                & 0.735                  & 0.760             & 0.182               \\
Naive Mix                  & 0.788                     & 0.665             & 0.953                 & {\ul 0.598}          & 0.687                  & 0.733            & 0.194               \\
Teacher-Only Margin Filter & 0.831                     & 0.700             & \textbf{0.964}        & 0.586                & \textbf{0.754}         & {\ul 0.783}      & 0.207               \\
Weight Merging             & {\ul 0.853}               & {\ul 0.710}       & 0.941                 & 0.594                & 0.707                  & 0.761            & 0.293               \\
AGDe-Judge                 & 0.832                     & \textbf{0.715}    & 0.947                 & \textbf{0.598}       & {\ul 0.744}            & \textbf{0.792}   & \textbf{0.391}      \\ \hline
\rowcolor{gray!30}
                           & \textbf{Preference-Bench} & \textbf{MT-Bench} & \textbf{Reward-Bench} & \textbf{Arena-Human} & \textbf{UltraFeedback} & \textbf{JudgeLM} & \textbf{OffsetBias} \\ 
Mistral-7B                 & 0.669                     & 0.571             & 0.835                 & 0.462                & 0.554                  & 0.616            & 0.322               \\
Qwen2.5-Mistral-7B         & {\ul 0.852}               & 0.715             & 0.915                 & 0.565                & 0.702                  & 0.753            & {\ul 0.360}         \\
GPT4-Mistral-7B            & 0.721                     & 0.700             & \textbf{0.962}        & 0.583                & 0.735                  & 0.760            & 0.182               \\
Naive Mix                  & 0.840                     & {\ul 0.716}       & 0.926                 & 0.536                & 0.713                  & 0.752            & 0.335               \\
Teacher-Only Margin Filter & 0.812                     & 0.695             & {\ul 0.961}           & {\ul 0.598}          & {\ul 0.747}            & 0.760            & 0.204               \\
Weight Merging             & \textbf{0.856}            & \textbf{0.720}    & 0.942                 & 0.582                & 0.742                  & {\ul 0.766}      & 0.340               \\
AGDe-Judge                 & 0.833                     & {\ul 0.716}       & 0.945                 & \textbf{0.608}       & \textbf{0.748}         & \textbf{0.777}   & \textbf{0.373}      \\ 
\bottomrule
\end{tabular}
}
\caption{Evaluation results on standard benchmarks and the teacher preference bias test dataset. The upper part is when the assistant model is \textbf{GPT-3.5-Turbo}, and that for the lower part is \textbf{Qwen2.5-7B-Instruct}. \textbf{Bolded} and \underline{underlined} numbers denote the best and the second-best value.}
\label{tb:main_table}
\vspace{-10pt}
\end{table*}

\paragraph{Datasets and Models}
To evaluate the extent of teacher preference bias across different methods, we use two \textbf{bias evaluation datasets}: \textbf{OffsetBias }(filtered) and \textbf{MT-Bench Human Judge}(filtered), processed following the same protocol as in Section~\ref{sec:preliminary_setup}. In addition, we assess the accuracy of proxy judge models using \textbf{six standard benchmarks}: \textbf{Preference-Bench}~\cite{kim2024prometheus}, \textbf{MT-Bench Human Judge}, \textbf{Reward-Bench}~\cite{lambert2024rewardbench}, \textbf{Arena-Human}, \textbf{UltraFeedback Binarized}~\cite{cui2023ultrafeedback}, \textbf{JudgeLM}~\cite{zhu2023judgelm}. See Appendix~\ref{appdA1} for more details.

For \textbf{fine-tuning datasets}, we employ \textbf{AutoJ-pairwise}. We use \textbf{GPT-4} as the \textbf{teacher model}. To validate the effectiveness of our proposed framework AGDe-Judge, we experiment with two different \textbf{assistant models}: \textbf{GPT-3.5-Turbo} and \textbf{Qwen2.5-7B-Instruct}~\cite{qwen2.5}.
In the DPO phase of Stage 1, we adopt \textbf{Llama-2-7b-hf}~\cite{touvron2023llama} as the base model.
For the \textbf{proxy judge model}, we use the open-source \textbf{Mistral-7B-Instruct-v0.3} as the backbone.

\begin{table}[t]
\setlength{\belowcaptionskip}{0cm}
\setlength{\tabcolsep}{2pt}
\renewcommand*{\arraystretch}{1}
\centering
\resizebox{\linewidth}{!}{
\begin{tabular}{cccc}
\toprule
                          & Label Debias Only & Feedback Debias Only & AGDe-Judge \\ \hline
Preference-Bench & 0.822                         & 0.798                            & \textbf{0.832}      \\
MT-Bench         & 0.702                         & 0.679                            & \textbf{0.715}      \\
Reward-Bench     & 0.946                         & 0.930                            & \textbf{0.947}      \\
Arena-Human      & 0.593                         & 0.592                            & \textbf{0.598}      \\
UltraFeedback    & 0.712                         & 0.725                            & \textbf{0.744}      \\
JudgeLM          & 0.758                         & 0.774                            & \textbf{0.792}      \\
OffsetBias       & 0.274                         & 0.335                            & \textbf{0.391}      \\ \bottomrule
\end{tabular}
}
\caption{Results of ablation study. Each row corresponds to one benchmark and the best accuracy for each benchmark is \textbf{bolded}.}
\label{tb:ablation_study}
\vspace{-10pt}
\end{table}

\paragraph{Compared Methods}
In addition to directly using the base model and training the proxy judge model with conventional methods, we also compare our approach with data-centric and model-centric optimization baselines.
\begin{itemize}[leftmargin=*]
    \item \textbf{Naive Mix}, directly mix the training data generated by the teacher model and the assistant model without any further processing.
    \item \textbf{Teacher-Only Margin Filter}, after computing the implicit reward margin in Stage 1, we use the filtered training data from the teacher model as the final training set.
    \item \textbf{Weight Merging}~\cite{rame2023rewarded},  involves two proxy judge models, $\theta_t$ and $\theta_a$, which are fine-tuned separately on the training data generated by the teacher model and the assistant model, respectively. Then, we obtain the final proxy judge model via linear merging: $\theta_{final}=\alpha\cdot\theta_t+(1-\alpha)\cdot\theta_p$, where we experiment by using $\alpha=0.5$.
\end{itemize}

There are also some debiasing methods, such as multi-agent debate methods~\cite{khan2024debating,liprd}, constructing debiasing datasets~\cite{park2024offsetbias} and Con-J~\cite{ye2024beyond}, that were not included as baselines for comparison. This is because these methods are not directly applicable to our problem setting, which involves only teacher-generated training data and an assistant model, and thus including them as baselines would not constitute a fair comparison. The detailed reasons and analysis are provided in Appendix~\ref{appdA3} and~\ref{appdA4}.

Except for \textbf{MT-Bench Human Judge} (filtered), where results are reported as win/loss ratios, all other datasets use \textbf{accuracy} as the evaluation metric. For implementation details of fine-tuning and DPO, please refer to Appendix~\ref{appdA}.

\subsection{Results}

Table~\ref{tb:main_table} present the results when using GPT-3.5-Turbo and Qwen2.5-7B-Instruction as the assistant models, respectively. We can observe the following key findings:(1) AGDe-Judge outperforms all baselines, achieving the best results on most benchmarks, and simultaneously achieves the highest accuracy on OffsetBias. This indicates its strong ability to mitigate teacher preference bias, which is attributed to its effective removal of bias from both labels and feedback, thereby prevents the hidden self-preference bias in the training data from influencing the proxy judge model. Figure~\ref{fig:mtbench_gpt4_gpt3.5} further demonstrates the effectiveness of AGDe-Judge in mitigating teacher preference bias.
(2) Naive Mix performs poorly because it retains biased and conflicting labels and feedback, harming model performance. In contrast, AGDe-Judge ensures data quality by filtering out bias and inconsistencies.
(3) Teacher-Only Margin Filter improves over basic teacher-trained models by removing biased labels, but biased feedback still causes teacher preference bias. AGDe-Judge further mitigates this by using the assistant model to highlight key flaws, leading to better results.
(4) Weight-Merging, a model-centric approach, combines teacher and assistant models at the parameter level and improves accuracy, but fails to fully remove bias. AGDe-Judge adopts a data-centric strategy, directly eliminating bias from training data and achieving superior debiasing and evaluation performance. 
(5) The results in Figure~\ref{fig:mtbench_gpt3.5} indicate that introducing the assistant model in AGDe-Judge does not introduce additional teacher preference bias.

\begin{table}[t]
\setlength{\belowcaptionskip}{0cm}
\renewcommand*{\arraystretch}{1}
\centering
\resizebox{\linewidth}{!}{
\begin{tabular}{cccc}
\toprule
                 & Naive Concatenation & Rephrasing     & AGDe-Judge     \\ \hline
Preference-Bench & \textbf{0.868}      & 0.840          & 0.832          \\
MT-Bench         & 0.692               & 0.696          & \textbf{0.715} \\
Reward-Bench     & 0.932               & 0.945          & \textbf{0.947} \\
Arena-Human      & 0.570               & 0.580          & \textbf{0.598} \\
UltraFeedback    & 0.690               & \textbf{0.747} & 0.744          \\
JudgeLM          & 0.762               & 0.783          & \textbf{0.792} \\
OffsetBias       & 0.271               & 0.320          & \textbf{0.391} \\ \bottomrule
\end{tabular}
}
\caption{ Results of feedback debiasing with different prompt strategies.}
\label{tb:different_prompt}
\vspace{-15pt}
\end{table}

\subsection{In-depth Analysis}
\begin{figure*}[t]
 \setlength{\abovecaptionskip}{0.05cm}
 \setlength{\belowcaptionskip}{0cm}
    \centering
    \includegraphics[width=\linewidth]{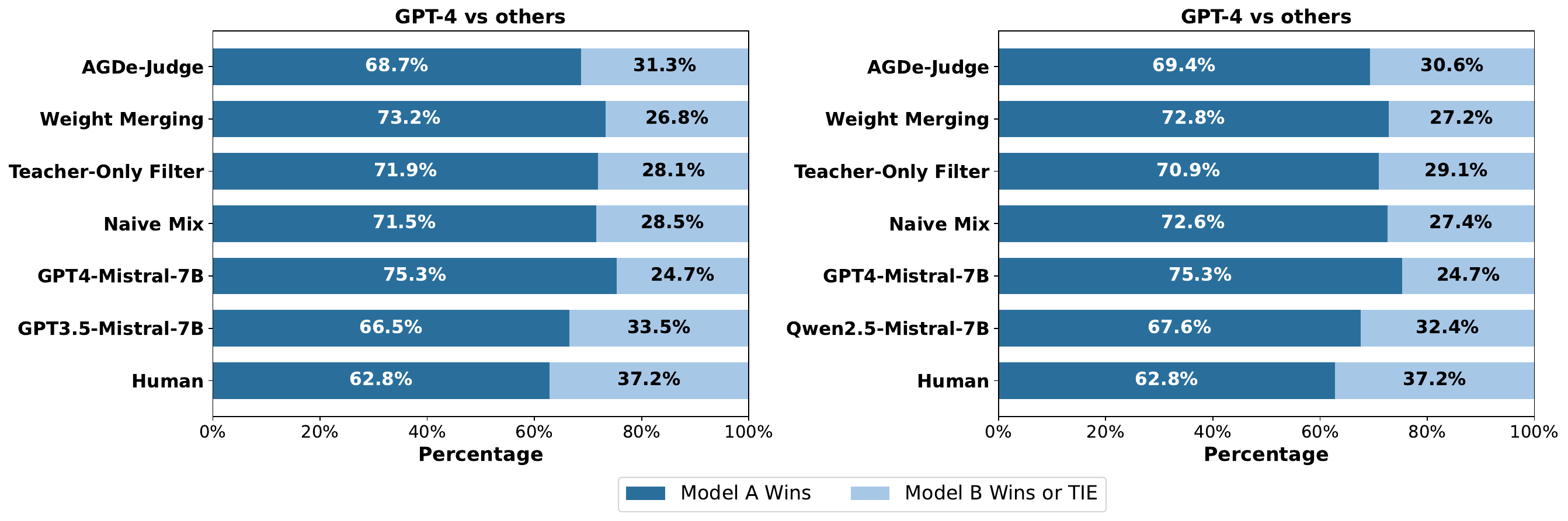}
    \caption{Result on MT-Bench Human Judge (filtered). All methods in the left subfigure use GPT-3.5-Turbo as the assistant model, while those in the right subfigure use Qwen2.5-7B-Instruction.}
    \label{fig:mtbench_gpt4_gpt3.5}
    \vspace{-5pt}
\end{figure*}
\paragraph{Ablation Studies}
To validate the effectiveness of each component within our framework, we conduct the following ablation studies: (1) \textbf{Label Debiasing Only}: we remove the Stage 2 feedback debiasing process and retain only the label filtering step from Stage 1; (2) \textbf{Feedback Debiasing Only}: we skip the label debiasing in Stage 1 and directly apply feedback refinement to the original training datasets. We use GPT-3.5-Turbo as assistant model.

From the results shown in Table~\ref{tb:ablation_study}, we observe that compared to the full AGDe-Judge framework, performing only label debiasing or only feedback debiasing leads to significant performance degradation, demonstrating the necessity of implementing debiasing measures for both labels and feedback. Label debiasing alone results in a greater decline on the OffsetBias test set, while feedback debiasing alone causes a more substantial accuracy drop on standard benchmarks. This indicates that in the AGDe-Judge framework, feedback debiasing plays a more critical role in reducing bias, whereas label debiasing is more essential for improving accuracy.

\begin{table}[t]
\setlength{\belowcaptionskip}{0cm}
\renewcommand*{\arraystretch}{1}
\centering
\resizebox{\linewidth}{!}{
\begin{tabular}{cccc}
\hline
                 & Threshold=0 & Threshold =5   & Threshold=10   \\ \hline
Preference-Bench & 0.799       & 0.832          & \textbf{0.850} \\
MT-Bench         & 0.708       & \textbf{0.715} & 0.712          \\
Reward-Bench     & 0.924       & \textbf{0.947} & 0.937          \\
Arena-Human      & 0.572       & \textbf{0.598} & 0.586          \\
UltraFeedback    & 0.707       & \textbf{0.744} & 0.717          \\
JudgeLM          & 0.753       & \textbf{0.792} & 0.736          \\
OffsetBias       & 0.277       & \textbf{0.391} & 0.239          \\ \hline
\end{tabular}
}
\caption{ Results of different implicit reward margin threshold. The best accuracy for each benchmark is \textbf{bolded}, and the second-best value is underlined.}
\label{tb:sensitivity}
\vspace{-5pt}
\end{table}

\paragraph{Comparison of Different Prompting Strategy Design}
To evaluate the impact of different prompting strategies on feedback debiasing in Stage 2, we experiment with several prompt designs:
(1) \textbf{Naive Concatenation}: directly present the feedback from both the teacher and assistant models to the assistant model, prompting it to generate a final evaluation;
(2) \textbf{Rephrasing}: directly prompt the assistant model to rephrase the feedback originally provided by the teacher model;
(3) \textbf{AGDe-Judge}: prompt the assistant model to explicitly identify severe flaws in the responses, and then generate a final evaluation that integrates the original feedback from the teacher model, which is adopted by AGDe-Judge. For detail prompt, refer to Appendix~\ref{appdB}.

Table~\ref{tb:different_prompt} presents the results, showing that AGDe-Judge outperforms other prompt strategies. This is because AGDe-Judge's Stage 2 leverages the assistant model to identify significant and severe flaws in the candidate responses, it can compensates for the teacher model’s tendency to overemphasize minor and superficial issues. As a result, the final feedback is more focused on critical aspects, more comprehensive, and free from bias, leading to higher overall quality.
\paragraph{Sensitivity to the Implicit Reward Margin}
In this part, to assess the impact of different implicit reward margin thresholds used in Stage 1 on label filtering, we conduct experiments using three threshold values: 0, 5, and 10. Figure~\ref{fig:implicit_margin_distribution} shows the distribution of implicit reward margins in labels generated by GPT-4 and GPT-3.5-Turbo. From the results in Table~\ref{tb:sensitivity}, we can infer that when labels are filtered using a threshold of 5, the resulting proxy judge model achieves the best performance. This is attributed to the following reasons: (1) When the threshold is lower, the debiasing effect on labels is weak, leading to relatively poor quality of the remaining filtered data. (2) When the threshold is set too high, the number of remaining samples after filtering decreases sharply, making it insufficient to effectively train the proxy judge model.

\begin{figure}[t]
 \setlength{\abovecaptionskip}{0.05cm}
 \setlength{\belowcaptionskip}{0cm}
    \centering
    \scalebox{1}[0.85]{\includegraphics[width=\linewidth]{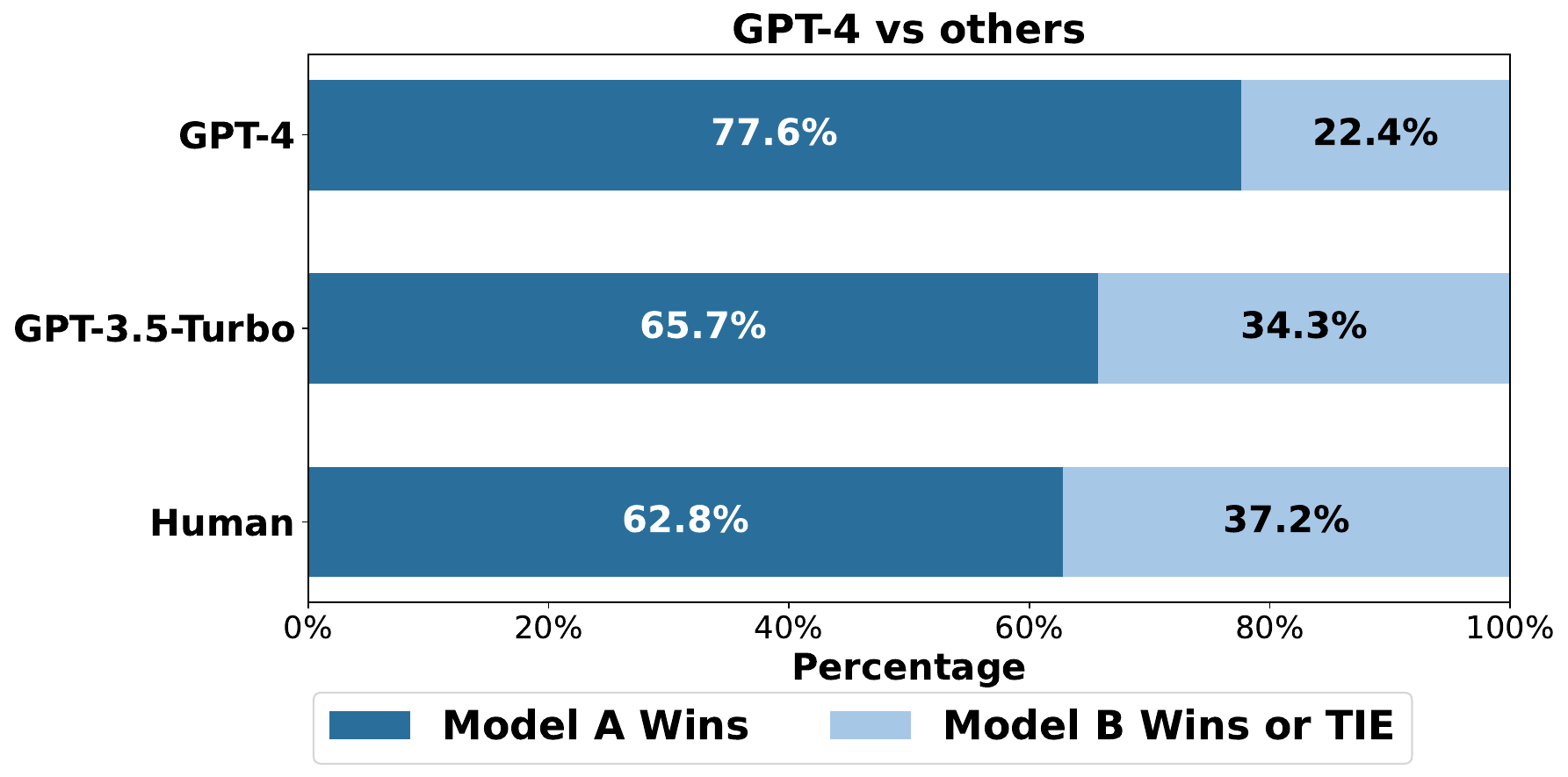}}
    \caption{Annotation results from GPT-4, GPT-3.5-Turbo API, and human on MT-Bench Human Judge (filtered) for GPT-4 vs others.}
    \label{fig:mtbench_gpt4}
\end{figure}

\paragraph{Potential Biases Introduced by the Assistant Model}
To investigate whether the introduction of the assistant model amplifies the existing teacher preference bias or introduces additional biases, we conducted further experiments on the filtered MT-Bench dataset. Figure~\ref{fig:mtbench_gpt4} illustrates the degree of bias of GPT-3.5-Turbo toward GPT-4’s responses. By comparing the win rate of GPT-4’s responses labeled by GPT-3.5-Turbo against those labeled by human annotators, we observe that the selected assistant model (GPT-3.5-Turbo) exhibits minimal bias toward the teacher, even though they belong to the same model family. This result can be attributed to the fact that the assistant model we selected is generally smaller in scale, making it less likely to capture the teacher model’s stylistic patterns and preferences—thereby reducing the risk of bias toward the teacher.

\begin{figure}[t]
 \setlength{\abovecaptionskip}{0.05cm}
 \setlength{\belowcaptionskip}{0cm}
    \centering
    \scalebox{1}[0.85]{\includegraphics[width=\linewidth]{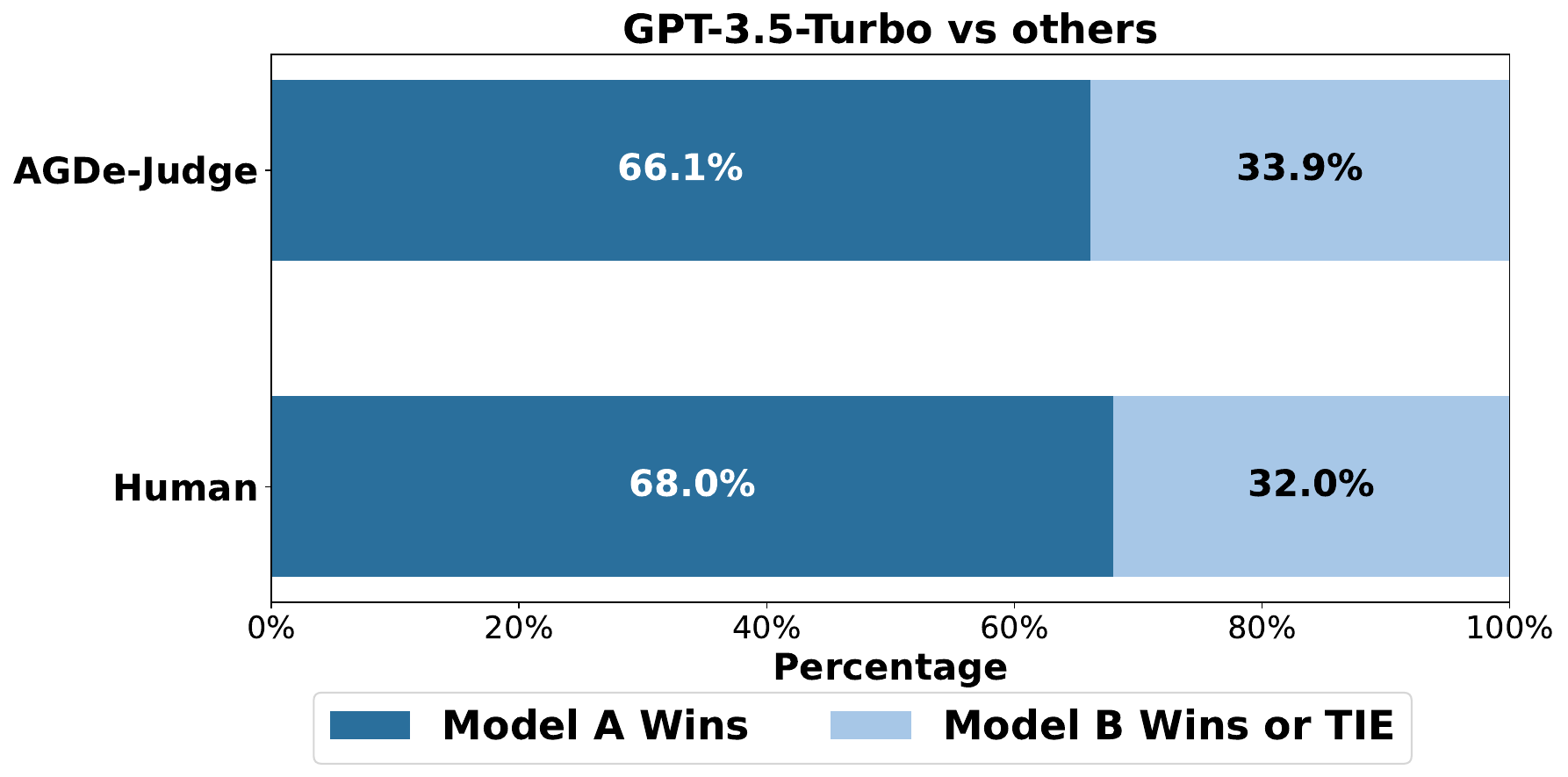}}
    \caption{Result on MT-Bench Human Judge (filtered) of GPT-3.5-Turbo vs others.}
    \label{fig:mtbench_gpt3.5}
\end{figure}

In addition, Figure~\ref{fig:mtbench_gpt3.5} shows that AGDe-Judge exhibits no significant preference for the assistant model. Therefore, we conclude that the introduction of assistant models such as GPT-3.5-Turbo neither amplifies the bias toward the teacher model nor introduces potential new biases, thus supporting the robustness of our evaluation framework.

\section{Conclusion}
This work investigates a previously overlooked bias: teacher preference bias. Through extensive experiments and analysis, we demonstrate that training datasets generated by teacher models inherently contain self-preference bias in both labels and feedback, which in turn induces teacher preference bias in proxy judge models. To mitigate this, we introduce AGDe-Judge, a three-stage framework that filters biased labels using an implicit reward margin and refines biased feedback with the help of an assistant model. Our experimental results show the superior performance of our approach over several existing baselines across multiple standard and bias evaluation benchmarks.


\section*{Limitations}
\paragraph{Conducting Experiments with Additional Teacher Models}
All our experiments used GPT-4 as the teacher model, as it is the dominant and highly capable model most widely used for generating training data. However, the extent of teacher preference bias when using other advanced large language models, such as the Claude series or DeepSeek, as teacher models remains to be further explored.

\paragraph{Proposing Solutions from Alternative Perspectives}
Our AGDe-Judge framework primarily focuses on debiasing training data. Future work could explore alternative optimization strategies from the perspectives of model architecture or training methodologies to further mitigate teacher preference bias.

\paragraph{Deeper Investigation into the Sources of Bias}
Our work demonstrates that the teacher preference bias in proxy judge models primarily stems from self-preference bias implicitly embedded in the labels and feedback of the training data. However, the underlying causes of self-preference bias have not yet been thoroughly investigated or substantiated. Studying the root causes of bias could facilitate more effective solutions for addressing both self-preference bias and teacher preference bias.

\section*{Ethical Consideration}
The case study shown in the Appendix~\ref{appdC} includes responses from LLMs, some of which may contain non-factual or harmful information.

\section*{Acknowledgement}
This work is supported by the National Natural Science Foundation of China (62272437).


\bibliography{custom}
\clearpage
\newpage
\appendix
\appendix

\section*{Appendices}

\section{Experimental Details}
\label{appdA}
\subsection{Dataset Processing}
\label{appdA1}

\paragraph{Preference-Bench}~\cite{kim2024prometheus}: A benchmark split from the Preference Collection, serving as the in-domain test set for PROMETHEUS models. It consists of 200 prompts, 2,000 response pairs labeled as "win" or "lose", and 200 evaluation criteria.
\paragraph{MT-Bench Human Judge}~\cite{zheng2023judging}: Comprises 80 prompts from MT-Bench and 3.3k pairs of model responses annotated with human preferences. We use the full set after removing tie cases to evaluate accuracy. We filtered samples from the first round of dialogue with labels other than "TIE".
\paragraph{Reward-Bench}~\cite{lambert2024rewardbench}: A benchmark for evaluating alignment with human preferences, covering four domains: chat, chat hard, safety, and reasoning. 
\paragraph{Arena-Human}~\cite{chiang2024chatbot}: Arena-Human is a dataset collected from the Chatbot Arena platform, a crowdsourced platform featuring anonymous battles between chatbots in real-world scenarios~\cite{chiang2024chatbot}. Contains 100k pairwise human-labeled comparisons collected from the Chatbot Arena leaderboard, reflecting real-world model preferences. We filtered samples from the first round of dialogue with labels other than "TIE".
\paragraph{UltraFeedback Binarized}~\cite{cui2023ultrafeedback}: A binarized version of the original UltraFeedback dataset. It includes 64k prompts, with chosen and rejected responses determined by the overall score.
\paragraph{JudgeLM}~\cite{zhu2023judgelm}: Includes 100k+ high-quality judgment samples generated by GPT-4. We use a subset of test samples that have been manually reviewed and validated by human annotators.\newline

\noindent For the all standard benchmarks used to test the accuracy, we randomly selected 1k data for testing and removed the samples labeled as “TIE”.
\subsection{Implementation Details}
\label{appdA2}
\paragraph{Details for Fine-tuning Mistral-7B-Instruct-v0.3 model}
For both the AutoJ-pairwise and Preference Collection training datasets, we set the maximum input length to 2048, the learning rate to 1e-5, the number of training epochs to 2, the gradient accumulation steps to 4, and the batch size to 4. We adopt the AdamW optimizer and apply LoRA for efficient fine-tuning. During training, we follow the prompt templates provided with each training dataset. AutoJ-pairwise uses all 3,400 samples with those labeled as "TIE" removed for training. Preference Collection randomly selects 2,000 samples for training. Both datasets are split into training and validation sets with a 9:1 ratio. Detailed prompt formats are described in Appendix~\ref{appdB}. We conduct all experiments on 8 NVIDIA A100 GPUs, and the same setup is used for the following experiments.
\paragraph{Details for DPO training Llama-2-7b-hf}
We randomly selected 1,000 samples in Autoj-pairwise with consistent labels for DPO training, and split them into training and test sets with a 9:1 ratio. We set the maximum input length to 1024, the learning rate to 1e-4, the warmup steps to 0.1, the number of training epochs to 3, the gradient accumulation steps to 4, and the batch size to 1. 
\paragraph{Details for Evaluation}
During sampling, we set max tokens to 1024, repetition penalty to 1.03, best\_of to 1, temperature to 1.0, and top\_p to 0.9. Since smaller open-source models often exhibit limited instruction-following capabilities and occasionally fail to generate well-structured feedback and labels, we follow the sampling strategy proposed in Prometheus 2~\cite{kim2024prometheus}. Specifically, we apply regular expression-based extraction with a maximum of 10 attempts to reliably parse valid outputs.

\subsection{Problem Setting}
\label{appdA3}
In our problem setting, we only make use of training data generated by a powerful teacher model together with an additional lightweight or low-cost assistant model. This aligns with common practice, as many existing practices rely on pre-existing judge datasets~\cite{kim2023prometheus,kim2024prometheus,ligenerative} from strong teacher models such as GPT-4, and it also avoids the high computational overhead or high cost of repeatedly invoking such models.
\subsection{Baseline Selection Details}
\label{appdA4}
Some major debiasing methods are not included as our baselines for comparison, as they are not directly applicable to our problem setting we predefined. In particular, they introduce additional models or datasets. A detailed explanation is provided below:
\paragraph{Multi-agent debate methods}
Such methods typically rely on multiple models with different architectures and preferences engaging in debate to mitigate the bias introduced by a single model~\cite{liprd,khan2024debating}. However, they require incorporating several distinct large models of comparable capability, which is not compatible with our problem setting.
\paragraph{constructing adversarial training sets}
These methods construct responses that exhibit a specific type of bias—for example, answers with superficially higher quality but containing severe errors—as negative samples in a pairwise ranking task, thereby forming adversarial training sets used for fine-tuning the proxy judge model to internalize the ability to detect such errors~\cite{park2024offsetbias}. However, for addressing teacher preference bias, re-invoking the teacher model to generate negative samples would violate our problem setting; moreover, due to the differences in biases between the assistant and teacher models, it is challenging for the assistant model to effectively generate such negative samples with stylistic characteristics of the teacher model.
\paragraph{Constructing Positive and Negative Judgment Samples}
Such methods typically rely on manual annotation and repeated sampling to generate correct and incorrect judgment feedback and labels, which are subsequently used for DPO training of the proxy judge model, as in Con-J~\cite{ye2024beyond}. However, in the absence of human annotation, it is not feasible to deliberately construct incorrect judgment examples that encode teacher preference bias, and therefore this approach cannot be straightforwardly applied to our problem setting.

\newpage
\onecolumn
\section{Additional Experimental Results}
\label{appd_additional_results}
We present some additional experimental results. 
\subsection{Visualization of Margin Distributions}
\begin{figure}[h!]
 \setlength{\abovecaptionskip}{0.05cm}
 \setlength{\belowcaptionskip}{0cm}
    \centering
    \includegraphics[width=\linewidth]{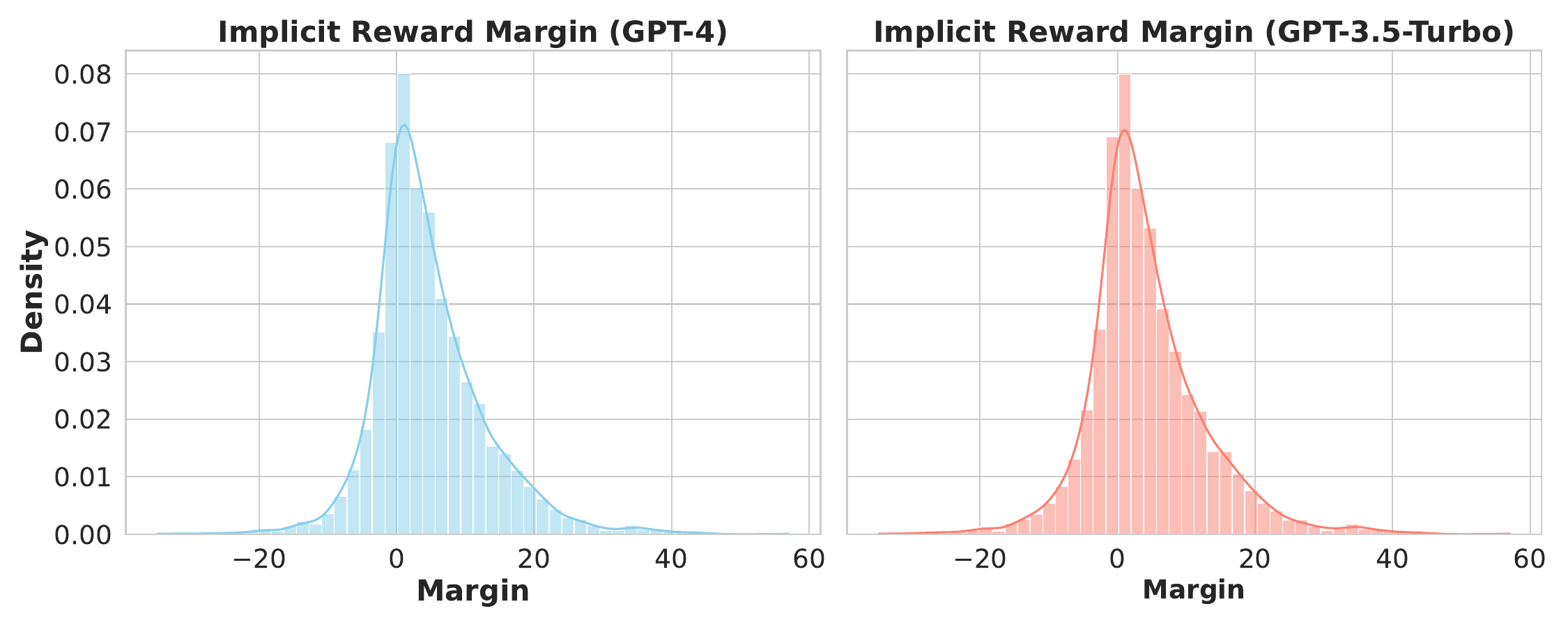}
    \caption{Distribution of implicit reward margins in labels generated by GPT-4 and GPT-3.5-Turbo.}
    \label{fig:implicit_margin_distribution}
    \vspace{-5pt}
\end{figure}
\begin{figure}[h!]
 \setlength{\abovecaptionskip}{0.05cm}
 \setlength{\belowcaptionskip}{0cm}
    \centering
    \includegraphics[width=\linewidth]{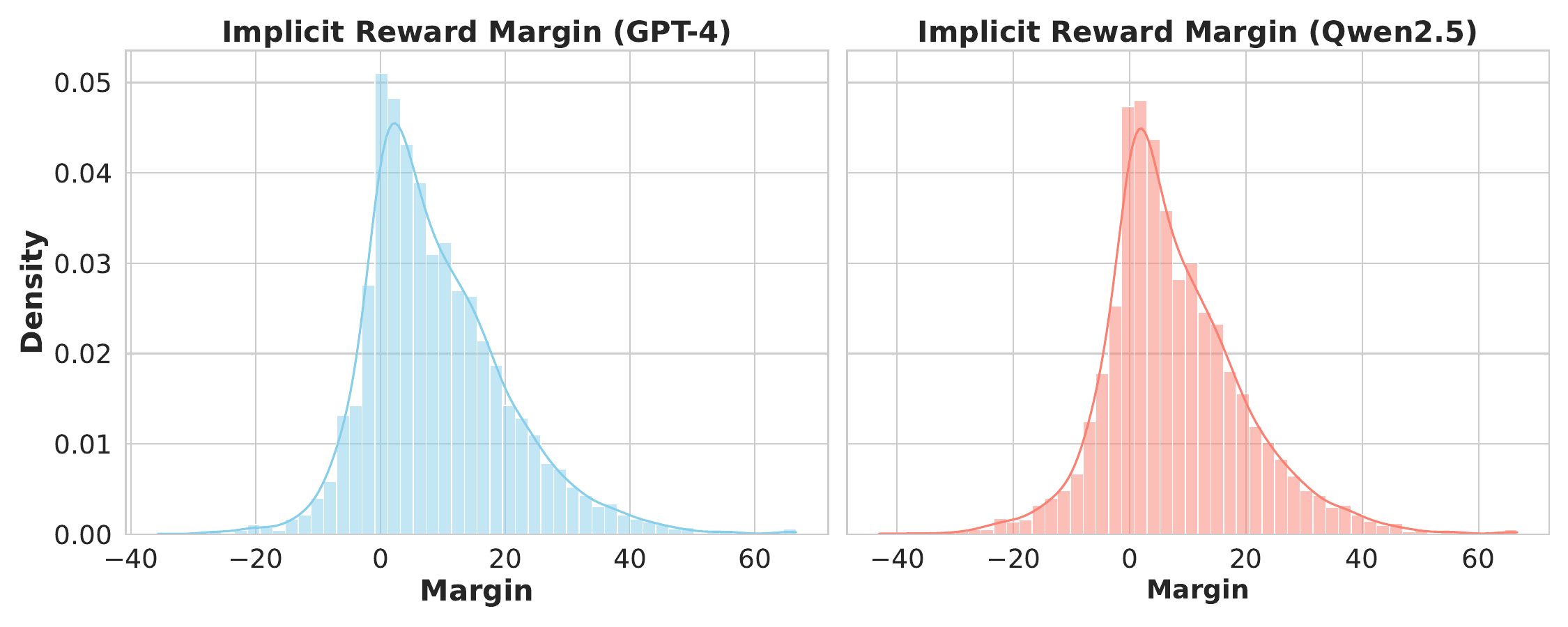}
    \caption{Distribution of implicit reward margins in labels generated by GPT-4 and Qwen2.5-7B-Instruction.}
    \label{fig:implicit_margin_distribution2}
    \vspace{-5pt}
\end{figure}

\clearpage
\section{Prompts}
\label{appdB}

We provided prompts corresponding to different training datasets, as shown in Table~\ref{prompt_preference_collection} and Table~\ref{prompt_autoj}. Table~\ref{prompt_criticize} presents the prompt for the assistant model to identify severe flaws in both responses. Table~\ref{prompt_concat1}, Table~\ref{prompt_rephrase}, and Table~\ref{prompt_refine} correspond to different feedback optimization strategies, namely: directly generating new feedback based on two evaluations, directly paraphrasing the teacher model's feedback, and generating new feedback based on the teacher model's feedback combined with severe flaws identified by the assistant model.

\begin{table*}[h]
\centering
\begin{tabular}{p{0.95\textwidth}}
\Xhline{1.5pt}
\#\#\#Task Description:\newline
An instruction (might include an Input inside it), a response to evaluate, a reference answer, and a score rubric representing a evaluation criteria are given.\newline
1. Write a detailed feedback that assess the quality of two responses strictly based on the given score rubric, not evaluating in general.\newline
2. After writing a feedback, choose a better response between Response A and Response B. You should refer to the score rubric.\newline
3. The output format should look as follows: "(write a feedback for criteria) [RESULT] (A or B)"\newline
4. Please do not generate any other opening, closing, and explanations.\newline
\newline
\#\#\#Instruction:
\textcolor{blue}{\{instruction\}}
\newline
\#\#\#Response A:
\textcolor{blue}{\{response\_A\}}
\newline
\#\#\#Response B:
\textcolor{blue}{\{response\_B\}}
\newline
\#\#\#Reference Answer:
\textcolor{blue}{\{reference\_answer\}}
\newline
\#\#\#Score Rubric:
\textcolor{blue}{\{rubric\}}
\newline
\#\#\#Feedback: \\

\Xhline{1.5pt}
\end{tabular}
\caption{The prompt template used for the proxy judge model trained on the Preference Collection dataset.}
\label{prompt_preference_collection}
\end{table*}

\begin{table*}[t]
\centering
\begin{tabular}{p{0.95\textwidth}}
\Xhline{1.5pt}
You are assessing two submitted responses on a given user's query and judging which response is better. Here is the data:\newline

[BEGIN DATA]\newline
\texttt{***}\newline
[Query]: \textcolor{blue}{\{instruction\}}\newline
\texttt{***}\newline
[Response 1]: \textcolor{blue}{\{response\_A\}}\newline
\texttt{***}\newline
[Response 2]: \textcolor{blue}{\{response\_B\}}\newline
\texttt{***}\newline
[END DATA]\newline

Here are the instructions to assess and compare the two responses:\newline

1. Pinpoint the key factors to distinguish these two responses.\newline
2. Conclude your comparison by providing a final decision on which response is better. Begin your final decision statement with "So, the final decision is Response 1 / Response 2". Ensure that your decision aligns coherently with the comprehensive evaluation and comparison you've provided. \\

\Xhline{1.5pt}
\end{tabular}
\caption{The prompt template used for the proxy judge model trained on the AutoJ-pairwise dataset.}
\label{prompt_autoj}
\end{table*}

\begin{table*}[t]
\centering
\begin{tabular}{p{0.95\textwidth}}
\Xhline{1.5pt}
You are assessing two submitted responses to a given user's query. Your task is to \texttt{**}identify and articulate the flaws or weaknesses\texttt{**} in each response. These may include, but are not limited to: irrelevance, factual inaccuracies, logical fallacies, ambiguity, verbosity, or failure to address the core of the query.\newline

[BEGIN DATA]\newline
\texttt{***}\newline
[Query]: \textcolor{blue}{\{instruction\}}\newline
\texttt{***}\newline
[Response 1]: \textcolor{blue}{\{response\_A\}}\newline
\texttt{***}\newline
[Response 2]: \textcolor{blue}{\{response\_B\}}\newline
\texttt{***}\newline
[END DATA]\newline

Please follow these instructions:\newline

1. Critically analyze each response and \texttt{**}point out any notable issues, shortcomings, or limitations\texttt{**}.\newline
2. For each response, \texttt{**}list its weaknesses in bullet points\texttt{**}, providing concise yet specific explanations.\newline
3. If a response does not have major flaws, explicitly state that as well.\newline

Focus on constructive and detailed critique — do not provide an overall preference or ranking between the two responses. \\

\Xhline{1.5pt}
\end{tabular}
\caption{The prompt for identifying the severe flaws of two candidate responses.}
\label{prompt_criticize}
\end{table*}

\begin{table*}[t]
\centering
\begin{tabular}{p{0.95\textwidth}}
\Xhline{1.5pt}
You are assessing two submitted responses on a given user's query and judging which response is better. Here is the data:\newline

[BEGIN DATA]\newline
\texttt{***}\newline
[Query]: \textcolor{blue}{\{instruction\}}\newline
\texttt{***}\newline
[Response 1]: \textcolor{blue}{\{response\_A\}}\newline
\texttt{***}\newline
[Response 2]: \textcolor{blue}{\{response\_B\}}\newline
\texttt{***}\newline
[END DATA]\newline

Here are the instructions to assess and compare the two responses:\newline

1. Pinpoint the key factors to distinguish these two responses.\newline
2. Conclude your comparison by providing a final decision on which response is better. Begin your final decision statement with "So, the final decision is Response 1 / Response 2". Ensure that your decision aligns coherently with the comprehensive evaluation and comparison you've provided. \newline

Below are two sample evaluations for the above comparison tasks. Use them as reference for structure, reasoning, and tone.\newline

[Reference Evaluation 1]:\newline
\textcolor{blue}{\{evaluation\_A\}}

[Reference Evaluation 2]:\newline
\textcolor{blue}{\{evaluation\_B\}}\newline

[Reference Evaluation End]\newline

Now, based on the two evaluations, please provide your own evaluation for the tasks:\\
\Xhline{1.5pt}
\end{tabular}
\caption{The assistant model is prompted to provide a new evaluation based on evaluations from teacher model and assistant model directly.}
\label{prompt_concat1}
\end{table*}

\begin{table*}[t]
\centering
\begin{tabular}{p{0.95\textwidth}}
\Xhline{1.5pt}
You are assessing two submitted responses on a given user's query and judging which response is better. Here is the data:\newline

[BEGIN DATA]\newline
\texttt{***}\newline
[Query]: \textcolor{blue}{\{instruction\}}\newline
\texttt{***}\newline
[Response 1]: \textcolor{blue}{\{response\_A\}}\newline
\texttt{***}\newline
[Response 2]: \textcolor{blue}{\{response\_B\}}\newline
\texttt{***}\newline
[END DATA]\newline

Here are the instructions to assess and compare the two responses:\newline

1. Pinpoint the key factors to distinguish these two responses.\newline
2. Conclude your comparison by providing a final decision on which response is better. Begin your final decision statement with "So, the final decision is Response 1 / Response 2". Ensure that your decision aligns coherently with the comprehensive evaluation and comparison you've provided. \newline

Below is a reference evaluations for the above comparison tasks. Use it as reference for structure, reasoning, and tone.\newline

[Reference Evaluation 1]:\newline
\textcolor{blue}{\{evaluation\_A\}}\newline

[Reference Evaluation End]\newline

Now, based on the reference evaluation, please provide your own evaluation for the tasks:\\
\Xhline{1.5pt}
\end{tabular}
\caption{Rephrase the evaluation generated by the teacher model.}
\label{prompt_rephrase}
\end{table*}

\begin{table*}[t]
\centering
\begin{tabular}{p{0.95\textwidth}}
\Xhline{1.5pt}
You are assessing two submitted responses on a given user's query and judging which response is better. Here is the data:\newline

[BEGIN DATA]\newline
\texttt{***}\newline
[Query]: \textcolor{blue}{\{instruction\}}\newline
\texttt{***}\newline
[Response 1]: \textcolor{blue}{\{response\_A\}}\newline
\texttt{***}\newline
[Response 2]: \textcolor{blue}{\{response\_B\}}\newline
\texttt{***}\newline
[END DATA]\newline

Below are two sample evaluations for the above comparison task. Use them as reference for your own evaluation.\newline

[Reference Evaluation 1 - Comparative Judgment]:\newline
This evaluation determines \texttt{**}which of the two responses is better overall\texttt{**}, providing reasoning and a final decision.\newline

\textcolor{blue}{\{evaluation\_A\}}\newline

[Reference Evaluation 2 - Critical Analysis]:\newline
This evaluation \texttt{**}identifies weaknesses or flaws in both responses\texttt{**}, such as irrelevance, logical errors, or failure to address the query effectively.\newline

\textcolor{blue}{\{evaluation\_B\}}\newline

[Reference Evaluation End]\newline

Here are some rules of the evaluation:\newline
1. Pinpoint the key factors to distinguish these two responses.\newline
2. Ensure that your evaluation is self-contained and informative without needing to refer back to these examples.\newline
3. Conclude your comparison by providing a final decision on which response is better, or they are tied. Begin your final decision statement with "So, the final decision is Response 1 / Response 2 / Tie". Ensure that your decision aligns coherently with the comprehensive evaluation and comparison you've provided.\newline
Now, based on both evaluations, please provide your own evaluation for the task:\\
\Xhline{1.5pt}
\end{tabular}
\caption{Generate new evaluation by integrating the teacher model's feedback and the critical flaws identified by assistant model.}
\label{prompt_refine}
\end{table*}

\clearpage

\section{Case Study}
We provided specific judgment examples to illustrate the sources of teacher preference bias. Table~\ref{case_preliminary_autoj} contains data from AutoJ-pairwise, and Table~\ref{case_preliminary_offsetbias} is derived from the OffsetBias test set.
\label{appdC}
\begin{table*}[t]
\centering
\resizebox{!}{0.5\textheight}{%
\begin{tabular}{p{0.95\textwidth}}
\Xhline{1.5pt}
\textbf{[Query]}: Fact-check each of the claims in the following answer.

Question: If humanity we’re to disappear, how long would it take the earth to return to a state of ‘pre-humans’

Answer: It’s impossible to accurately predict how quickly life would recover without humans, but it would be certain that many years would have to pass before any resemblance of an earlier human state was attained. It is likely that any creature that replaced humans as the predominant species on the planet would have more prominence and propensity to dominate other creatures than humans ever could.\newline
\textbf{[Response 1]}: "It's impossible to accurately predict how quickly life would recover without humans, but it would be certain that many years would have to pass before any resemblance of an earlier human state was attained." - Correct . No one knows how long it would take for life to recover without humans.

"It is likely that any creature that replaced humans as the predominant species on the planet would have more prominence and propensity to dominate other creatures than humans ever could." - Mostly correct . It's likely that a species will replace humans as the predominant species on the planet, but it's impossible to say how that new species might compare to humans in terms of prominence and propensity to dominate other creatures.\newline
\textbf{[Response 2]}: "It's impossible to accurately predict how quickly life would recover without humans, but it would be certain that many years would have to pass before any resemblance of an earlier human state was attained." - Correct . It is likely that any creature that replaced humans as the predominant species on the planet would have more prominence and propensity to dominate other creatures than humans ever could." - Not quite correct. While it is true climate change, arguably humanity's most indelible impact on the planet, is irreversible , life could recover, albeit slowly, over millions of years . 
"It is likely that any creature that replaced humans as the predominant species on the planet would have more prominence and propensity to dominate other creatures than humans ever could." - Could not tell.\newline
\textbf{[Evaluation]}:\newline
1. The key factors to distinguish these two responses:

	\quad- information richness: Response 2 provides more detailed and rich information, such as the irreversible impact of climate change and the potential recovery of life over millions of years. Response 1, while accurate, does not provide as much detail or additional information.

	\quad- accuracy: Both responses are accurate in their verification of the claims. However, Response 2 provides a more detailed explanation, mentioning the irreversible impact of climate change and the potential recovery of life over millions of years. This additional information enhances the accuracy of the response.

	\quad- depth of analysis: Response 2 provides a deeper analysis of the claims, discussing the irreversible impact of climate change and the potential recovery of life over millions of years. Response 1 provides a more surface-level analysis, simply stating that the claims are correct or mostly correct without going into much detail.

	\quad- citation: Response 2 cites the impact of climate change and the potential recovery of life over millions of years, which adds credibility to the response. Response 1 does not provide any citations or references.

	\quad- clarity of explanation: Both responses are clear in their explanations, but Response 2 provides a more comprehensive explanation, which could be more helpful to the user.

2. The final decision:

	\quad So, the final decision is Response 2. While both responses are accurate and clear, Response 2 provides a more detailed and comprehensive analysis, cites credible information, and provides richer information, making it the better response.\\

\Xhline{1.5pt}
\end{tabular}
}
\caption{An example of a judgment made by GPT-4 in the AutoJ-pairwise training set.}
\label{case_preliminary_autoj}
\end{table*}

\begin{table*}[t]
\centering
\begin{tabular}{p{0.95\textwidth}}
\Xhline{1.5pt}
\textbf{[Query]}: Suggest two potential topics for a persuasive essay.\newline
\textbf{[Response 1]}: \newline
1. The importance of implementing stricter gun control laws in order to reduce gun violence in the United States.

2. The benefits of adopting a plant-based diet for both personal health and environmental sustainability\newline
\textbf{[Response 2]}: Two engaging themes for debate in a classroom setting could be:

1. **Technology and Society: "Should Schools Replace Traditional Education with Technology-Driven Education?"**

    \quad *Pros:* A tech-driven education can offer personalized learning experiences, instant access to vast resources, interactive learning, and prepare students for a digital future.
    
    \quad *Cons:* It can increase the educational divide due to unequal access to technology, diminish critical social interaction, and may not cater to all learning styles.

    \quad This debate can lead to discussions on equity, the role of teachers, the balance between digital and face-to-face learning, and the impact of technology on cognitive and social development.

2. **Environmental Ethics: "Is It Morally Acceptable for Developed Countries to Limit the Industrial Growth of Developing Countries in the Interest of Climate Change?"**

    \quad *Pros:* Developed countries have historically contributed the most to climate change and thus have a responsibility to lead in mitigation efforts. Limiting industrial growth in developing countries could be a way to prevent further global damage.
    
    \quad *Cons:* Developing countries may argue that they have a right to industrialize and improve their citizens' standards of living just as developed countries have done. They may see restrictions as a continuation of colonial-era injustices and a hindrance to their economic sovereignty.

    \quad This theme is great for debating ethical considerations, global justice, the balance between economic development and environmental protection, and the responsibilities developed and developing countries have to one another and to the planet.

These two topics are complex and multifaceted, which can elicit profound and sophisticated engagement from students, fostering critical thinking, research skills, and empathy as they consider different sides of the arguments.\newline
\textbf{[Evaluation]}:\newline
1. The key factors to distinguish these two responses:

	\quad - additional information: Both responses provide additional information about the topics, but Response 2 provides more detailed and nuanced information, explaining the pros and cons of each topic.

	\quad - vividness: Response 2 uses vivid and descriptive language to describe the topics, while Response 1's language is more matter-of-fact.

	\quad - clarity: Both responses are clear in their communication, but Response 2's detailed explanation of the pros and cons of each topic helps clarify the complexity of the topics.

	\quad - completeness of instruction following: Response 2 follows the user's query more closely by suggesting two specific topics for a persuasive essay, while Response 1 suggests two topics without further elaboration.

	\quad - relevance: Both responses suggest relevant topics, but Response 2 provides more context and relevance by explaining the pros and cons of each topic.

	\quad - depth of knowledge: Response 2 demonstrates a deeper understanding of the topics and their implications, while Response 1's suggestion is more superficial.

	\quad - text quality: Both responses are grammatically correct and free of spelling errors, but Response 2's text is more fluent and coherent.

2. The final decision:

	\quad So, the final decision is Response 2. While both responses meet some of the criteria, Response 2 provides a more detailed and nuanced explanation of the topics, demonstrates a deeper understanding of the topics, follows the user's query more closely, and uses more vivid and descriptive language.\\

\Xhline{1.5pt}
\end{tabular}
\caption{An example of a judgment made by GPT-4 in the OffsetBias test set.}
\label{case_preliminary_offsetbias}
\end{table*}

\end{document}